\documentclass[11pt,a4paper]{article}
\usepackage[hyperref]{acl2018}
\usepackage{times}
\usepackage{latexsym}
\usepackage{graphicx}
\usepackage{url}
\usepackage{float}
\usepackage{subcaption}

\usepackage[normalem]{ulem}
\useunder{\uline}{\ul}{}

\newcounter{tbsnr}
\newenvironment{tbs}
{\addtocounter{tbsnr}{1}\par\bigskip\noindent\fbox{\thetbsnr}
\hspace*{\fill}\begin{minipage}{7cm}\tt}
{\end{minipage}\hspace*{\fill}\bigskip}

\aclfinalcopy 
\def\aclpaperid{806} 

\title{\textit{Some of} Them Can be Guessed!\\Exploring the Effect of Linguistic Context in Predicting Quantifiers}

\author{Sandro Pezzelle$^*$, Shane Steinert-Threlkeld$^\dagger$, Raffaella Bernardi$^*$$^\ddagger$, Jakub Szymanik$^\dagger$\\
	      $^*$CIMeC - Center for Mind/Brain Sciences, $^\ddagger$DISI, University of Trento\\
	      $^\dagger$ILLC - Institute for Logic, Language and Computation, University of Amsterdam\\
	      $^*${\tt sandro.pezzelle@unitn.it}, $^\dagger${\tt s.n.m.steinert-threlkeld@uva.nl},\\
	      $^*$$^\ddagger${\tt raffaella.bernardi@unitn.it}, $^\dagger${\tt j.k.szymanik@uva.nl}}

\date{}

\begin{document}
\maketitle
\begin{abstract}

We study the role of linguistic context in predicting quantifiers (`few', `all'). We collect crowdsourced data from human participants and test various models in a \emph{local} (single-sentence) and a \emph{global} context (multi-sentence) condition. Models significantly out-perform humans in the former setting and are only slightly better in the latter. While human performance improves with more linguistic context (especially on proportional quantifiers), model performance suffers. Models are very effective in exploiting lexical and morpho-syntactic patterns; humans are better at genuinely understanding the meaning of the (global) context.


\end{abstract}

\section{Introduction}
\label{sec:introduction}

A typical exercise used to evaluate a language learner is the cloze deletion test~\cite{oller1973cloze}. In it, a word is removed and the learner must replace it. This requires the ability to understand the context and the vocabulary in order to identify the correct word. Therefore, the larger the linguistic context, the easier the test becomes. It has been recently shown that higher-ability test takers rely more on global information, with lower-ability test takers focusing more on the local context, i.e. information contained in the words immediately surrounding the gap~\cite{McCrayBrunfaut}.

\begin{figure}[t]
\begin{center}
\includegraphics[scale=0.35]{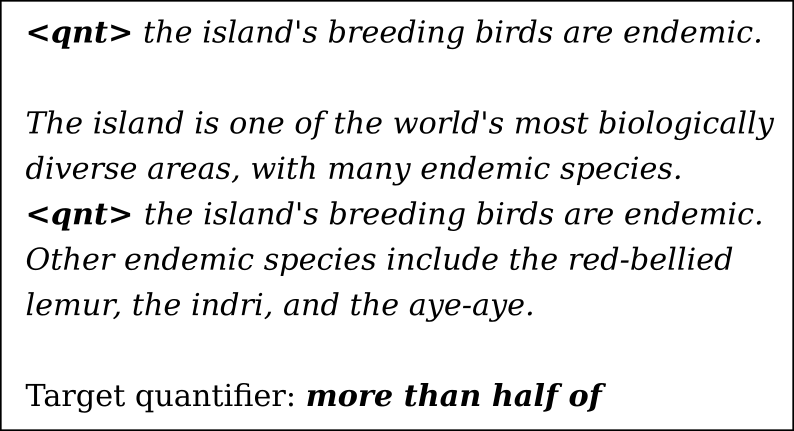}
\caption{Given a target sentence \emph{s\textsubscript{t}}, or \emph{s\textsubscript{t}} with the preceding and following sentence, the task is to predict the target quantifier replaced by \texttt{$<$qnt$>$}.} 
\label{ClozeTest}
\end{center}
\end{figure}

In this study, we explore the role of linguistic context in predicting generalized quantifiers (`few', `some', `most') in a cloze-test task (see Figure~\ref{ClozeTest}). Both human and model performance is evaluated in a \emph{local} (single-sentence) and a \emph{global} context (multi-sentence) condition to study the role of context and assess the cognitive plausibility of the models. The reasons we are interested in quantifiers are myriad. First, quantifiers are of central importance in linguistic semantics and its interface with cognitive science~\cite{Barwise1981,PW06,Szymanik:2015kq}. Second, the choice of quantifier depends both on local context (e.g., positive and negative quantifiers license different patterns of anaphoric reference) and global context (the degree of positivity/negativity is modulated by discourse specificity)~\cite{pate:quan09}. Third and more generally, the ability of predicting \textit{function words} in the cloze test represents a benchmark test for human linguistic competence~\cite{smith1971,hill:thego16}.

We conjecture that human performance will be boosted by more context and that this effect will be stronger for \emph{proportional} quantifiers (e.g. `few', `many', `most') than for \emph{logical} quantifiers (e.g. `none', `some', `all') because the former are more dependent on discourse context~\cite{Moxey:1993zr,Solt:2016aa}. In contrast, we expect models to be very effective in exploiting the local context~\cite{hill:thego16} but to suffer with a broader context, due to their reported inability to handle longer sequences~\cite{pape:thel16}. Both hypotheses are confirmed. The best models are very effective in the local context condition, where they significantly outperform humans. Moreover, model performance declines with more context, whereas human performance is boosted by the higher accuracy with proportional quantifiers like `many' and `most'. Finally, we show that best-performing models and humans make similar errors. In particuar, they tend to confound quantifiers that denote a similar `magnitude'~\cite{bass1974,newstead1987}.



Our contribution is twofold. First, we present a new task and results for training models to learn semantically-rich function words.\footnote{Data and code can be found at \texttt{github.com/sandropezzelle/fill-in-the-quant}} Second, we analyze the role of linguistic context in both humans and the models, with implications for cognitive plausibility and future modeling work.

\section{Datasets}
\label{sec:dataset}


To test our hypotheses, we need linguistic contexts containing quantifiers. To ensure similarity in the syntactic environment of the quantifiers, we focus on partitive uses: where the quantifier is followed by the preposition `of'. To avoid any effect of intensifiers like `very' and `so' and adverbs like `only' and `incredibly', we study only sentences in which the quantifier occurs at the beginning (see Figure~\ref{ClozeTest}). We experiment with a set of 9 quantifiers: `a few', `all', `almost all', `few', `many', `more than half', `most', `none', `some'. This set strikes the best trade-off between number of quantifiers and their frequency in our \textit{source} corpus, a large collection of written English including around 3B tokens.\footnote{A concatenation of BNC, ukWaC, and a 2009-dump of Wikipedia~\cite{baroni2014}.}

We build two datasets. One dataset -- \texttt{1-Sent} -- contains datapoints that only include the sentence with the quantifier (the \textit{target} sentence, \emph{s\textsubscript{t}}). The second -- \texttt{3-Sent} -- contains datapoints that are 3-sentence long: the target sentence (\emph{s\textsubscript{t}}) together with both the preceding (\emph{s\textsubscript{p}}) and following one (\emph{s\textsubscript{f}}). To directly analyze the effect of the linguistic context in the task, the target sentences are exactly the same in both settings. Indeed, \texttt{1-Sent} is obtained by simply extracting all target sentences \emph{$<$s\textsubscript{t}$>$} from \texttt{3-Sent} (\emph{$<$s\textsubscript{p}, s\textsubscript{t}, s\textsubscript{f}$>$}).

The \texttt{3-Sent} dataset is built as follows: (1) We split our source corpus into sentences and select those starting with a `\emph{quantifier} of' construction. Around 391K sentences of this type are found. (2) We tokenize the sentences and replace the quantifier at the beginning of the sentence (the \textit{target} quantifier) with the string \texttt{$<$qnt$>$}, to treat all target quantifiers as a single token. (3) We filter out sentences longer than 50 tokens (less than 6\% of the total), yielding around 369K sentences. (4) We select all cases for which both the preceding and the following sentence are at most 50-tokens long. We also ensure that the target quantifier does not occur again in the target sentence. (5) We ensure that each datapoint \emph{$<$s\textsubscript{p}, s\textsubscript{t}, s\textsubscript{f}$>$} is unique. The distribution of target quantifiers across the resulting 309K datapoints ranges from 1152 cases (`more than half') to 93801 cases (`some'). To keep the dataset balanced, we randomly select 1150 points for each quantifier, resulting in a dataset of 10350 datapoints. This was split into train (80\%), validation (10\%), and test (10\%) sets while keeping the balancing. Then, \texttt{1-Sent} is obtained by extracting the target sentences \emph{$<$s\textsubscript{t}$>$} from \emph{$<$s\textsubscript{p}, s\textsubscript{t}, s\textsubscript{f}$>$}.

\section{Human Evaluation}
\label{sec:humans}

\begin{table*}[t!]
\centering
\small
\begin{tabular}{|l|l|l|}
\hline
\textbf{type} & \textbf{text}                                                                                                                                                                                               & \textbf{quantifier}     \\ \hline
meaning        & \textit{$<$qnt$>$ the original station buildings survive \textbf{as they were used} as a source of materials\dots} & \textit{none of}           \\
PIs             & \textit{$<$qnt$>$ these stories have \textbf{ever} been substantiated.}                                                                                                                               & \textit{none of}           \\
contrast Q& \textit{$<$qnt$>$ the population died out, but \textbf{a} select \textbf{few} with the right kind of genetic instability\dots}                                            & \textit{most of}           \\
list           & \textit{$<$qnt$>$ their major research areas are \textbf{social inequality, group dynamics, social change}\dots}                             & \textit{some of}           \\
quantity       & \textit{$<$qnt$>$ those polled (\textbf{56\%}) said that they would be willing to pay for special events\dots}                                                    & \textit{more t. half of} \\
support Q   & \textit{$<$qnt$>$ you have found this to be the case - click here for \textbf{some of} customer comments.}                                                                                               & \textit{many of}           \\
lexicalized & \textit{$<$qnt$>$ \textbf{the time}, the interest rate is set on the lender's terms\dots}                                                                        & \textit{most of}           \\
syntax  & \textit{$<$qnt$>$ these events \textbf{was} serious.}                                                                                                                                                 & \textit{none of}           \\ \hline
\end{tabular}
\caption{Cues that might help human participants to predict the correct quantifier (\texttt{1-Sent}).}
\label{tab:clues}
\end{table*}

\subsection{Method}


We ran two crowdsourced experiments, one per condition. In both, native English speakers were asked to pick the correct quantifier to replace \texttt{$<$qnt$>$} after having carefully read and understood the surrounding linguistic context. When more than one quantifier sounds correct, participants were instructed to choose the one they think best for the context. To make the results of the two surveys directly comparable, the same randomly-sampled 506 datapoints from the validation sets are used. To avoid biasing responses, the 9 quantifiers were presented in alphabetical order. The surveys were carried out via CrowdFlower.\footnote{\texttt{https://www.figure-eight.com/}} Each participant was allowed to judge up to 25 points. To assess the judgments, 50 unambiguous cases per setting were manually selected by the native-English author and used as a benchmark. Overall, we collected judgments from 205 annotators in \texttt{1-Sent} (avg. 7.4 judgments/annotator) and from 116 in \texttt{3-Sent} (avg. 13.1). Accuracy is then computed by counting cases where at least 2 out of 3 annotators agree on the correct answer (i.e., inter-annotator agreement $\geq$ 0.67).




\subsection{Linguistic Analysis}\label{sec:linganalysis}

Overall, the task turns out to be easier in \texttt{3-Sent} (131/506 correctly-guessed cases; 0.258 accuracy) compared to \texttt{1-Sent} (112/506; 0.221 acc.). Broader linguistic context is thus generally beneficial to the task. To gain a better understanding of the results, we analyze the correctly-predicted cases and look for linguistic cues that might be helpful for carrying out the task. Table~\ref{tab:clues} reports examples from \texttt{1-Sent} for each of these cues.


We identify 8 main types of cues and manually annotate the cases accordingly. (1) \textbf{Meaning}: the quantifier can only be guessed by understanding and reasoning about the context; (2) \textbf{PIs}: Polarity Items like `ever', `never', `any' are licensed by specific quantifiers~\cite{krifka1995};
(3) \textbf{Contrast Q}: a contasting-magnitude quantifier embedded in an adversative clause;
(4) \textbf{Support Q}: a supporting-magnitude quantifier embedded in a coordinate or subordinate clause;
(5) \textbf{Quantity}: explicit quantitative information (numbers, percentages, fractions, etc.); (6) \textbf{Lexicalized}: lexicalized patterns like `\textit{most of} the time'; (7) \textbf{List}: the text immediately following the quantifier is a list introduced by verbs like `are' or `include'; (8) \textbf{Syntax}: morpho-syntactic cues, e.g. agreement.


\begin{figure*}[t!]
\centering
\begin{subfigure}{.505\textwidth}
 \centering
 \includegraphics[scale=0.175,keepaspectratio=true]{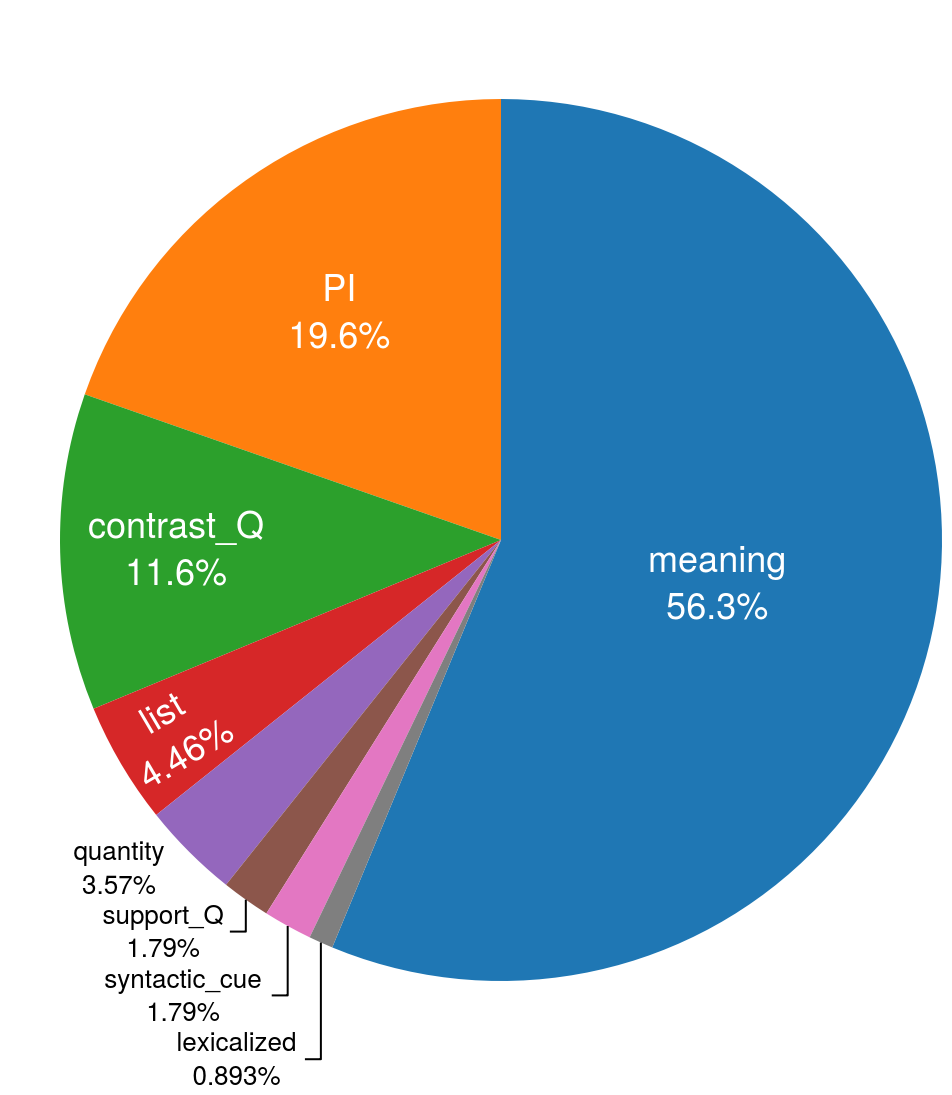}
\end{subfigure}%
\begin{subfigure}{.505\textwidth}
 \centering
 \includegraphics[scale=0.20,keepaspectratio=true]{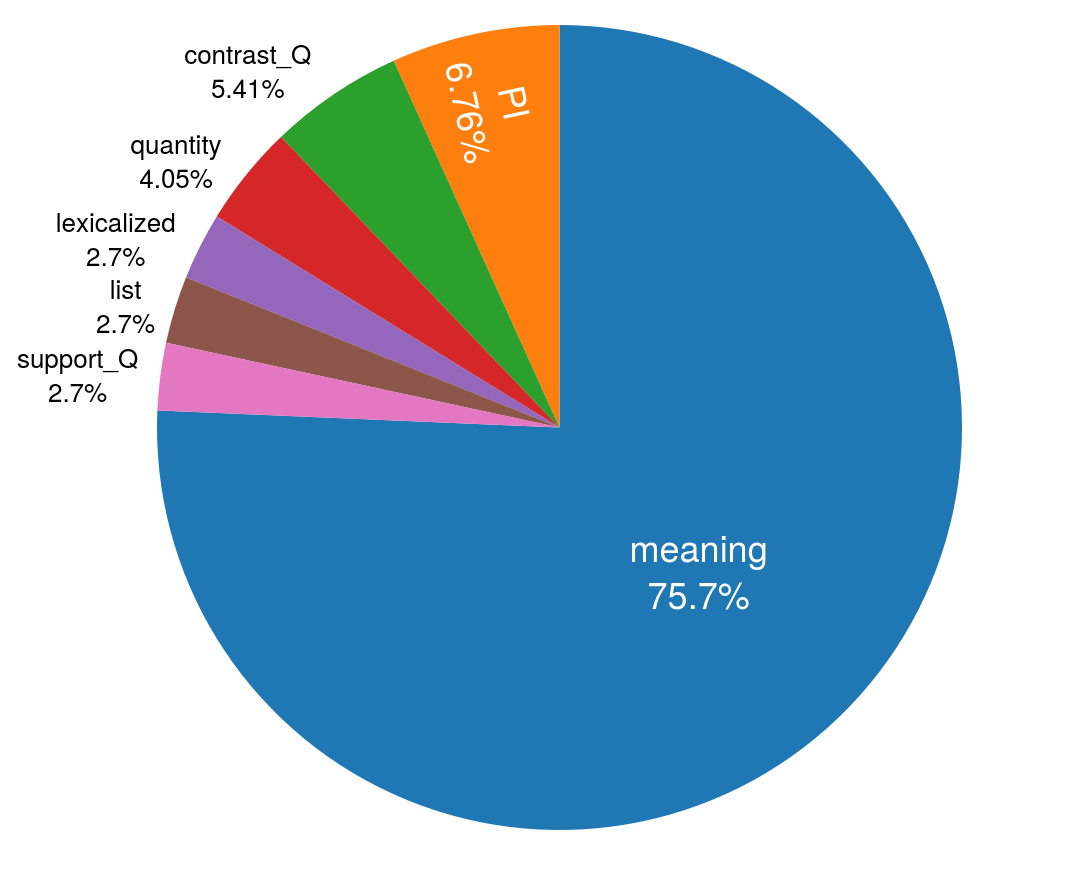}
\end{subfigure}
\caption{Left: Distribution of annotated cues across correcly-guessed cases in \texttt{1-Sent} (112 cases). Right: Distribution of cues across correctly-guessed cases in \texttt{3-Sent}, but not in \texttt{1-Sent} (74 cases).}\label{fig:pies}
\end{figure*}

Figure~\ref{fig:pies} (left) depicts the distribution of annotated cues in correctly-guessed cases of \texttt{1-Sent}. Around 44\% of these cases include cues besides meaning, suggesting that almost half of the cases can be possibly guessed by means of lexical factors such as PIs, quantity information, etc. As seen in Figure~\ref{fig:pies} (right), the role played by the meaning becomes much higher in \texttt{3-Sent}. Of the 74 cases that are correctly guessed in \texttt{3-Sent}, but not in \texttt{1-Sent}, more than 3 out of 4 do not display cues other than meaning. In the absence of lexical cues at the sentence level, the surrounding context thus plays a crucial role.

\section{Models}
\label{sec:experiments}


We test several models, that we briefly describe below. All models except \texttt{FastText} are implemented in Keras and use \texttt{ReLu} as activation function; they are trained for 50 epochs with categorical crossentropy, initialized with frozen 300-d \texttt{word2vec} embeddings~\cite{mikolov2013} pretrained on GoogleNews.\footnote{Available here: \texttt{http://bit.ly/1VxNC9t}} A thorough ablation study is carried out for each model to find the best configuration of parameters.\footnote{We experiment with all possible combinations obtained by varying (a) optimizer: \textit{adagrad}, \textit{adam}, \textit{nadam}; (b) hidden layers: 64 or 128 units; (c) dropout: 0.25, 0.5, 0.75.} The best configuration is chosen based on the lowest validation loss.


\paragraph{BoW-conc}

A bag-of-words (BoW) architecture which encodes a text as the \textit{concatenation} of the embeddings for each token. This representation is reduced by a hidden layer before softmax.


\paragraph{BoW-sum}

Same as above, but the text is encoded as the \textit{sum} of the embeddings.

\paragraph{FastText}

Simple network for text classification that has been shown to obtain performance comparable to deep learning models~\cite{fasttext}. \texttt{FastText} represents text as a hidden variable obtained by means of a BoW representation. 


\paragraph{CNN}

Simple Convolutional Neural Network (CNN) for text classification.\footnote{Adapted from: \texttt{http://bit.ly/2sFgOE1}} It has two convolutional layers (\texttt{Conv1D}) each followed by \texttt{MaxPooling}. A dense layer precedes softmax.

\paragraph{LSTM}

Standard Long-Short Term Memory network (LSTM)~\cite{lstm}. Variable-length sequences are padded with zeros to be as long as the maximum sequence in the dataset. To avoid taking into account cells padded with zero, the `mask zero' option is used.

\paragraph{bi-LSTM}

The Bidirectional LSTM~\cite{schuster1997} combines information from past and future states by duplicating the first recurrent layer and then combining the two hidden states. As above, padding and mask zero are used.


\paragraph{Att-LSTM}

LSTM augmented with an attention mechanism~\cite{Raffel2016}. A feed-forward neural network computes an importance weight for each hidden state of the LSTM; the weighted sum of the hidden states according to those weights is then fed into the final classifier.

\paragraph{AttCon-LSTM}

LSTM augmented with an attention mechanism using a learned \emph{context} vector~\cite{Yang2016a}. LSTM states are weighted by cosine similarity to the context vector. 

\begin{table}[b!]
\centering
\small
\begin{tabular}{|l|ll|ll|}
\hline
                & \multicolumn{2}{c|}{\texttt{1-Sent}}                                  & \multicolumn{2}{c|}{\texttt{3-Sent}}                                  \\ \cline{2-5} 
                & \multicolumn{1}{c}{\textit{val}} & \multicolumn{1}{c|}{\textit{test}} & \multicolumn{1}{c}{\textit{val}} & \multicolumn{1}{c|}{\textit{test}} \\ \hline\hline
\textit{chance} & 0.111                 & 0.111                     & 0.111                   & 0.111                     \\ \hline
BoW-conc        & 0.270                                 &  0.238                                  & 0.224                                 & 0.207                                   \\
BoW-sum         & 0.308                                 &  0.290                                  & 0.267                                  & 0.245                                   \\
fastText        & 0.305                            & 0.271                              & 0.297                            & 0.245                              \\ \hline
CNN             & 0.310                            & 0.304                                   & \textbf{0.298}                            & 0.257                                   \\ \hline
LSTM            & 0.315                            & 0.310                                   & 0.277                            & 0.253                                   \\
bi-LSTM         & 0.341                            & \textbf{0.337}                              & 0.279                            &  0.265                                  \\
Att-LSTM        & 0.319                                 & 0.324                                    & 0.287                                 & \textbf{0.291}                                   \\
AttCon-LSTM     & \textbf{0.343}                                 & 0.319                                   & 0.274                                 & 0.288                                   \\ \hline\hline
Humans & 0.221{*} & ------ & 0.258{*} & ------ \\ \hline
\end{tabular}
\caption{Accuracy of models and humans. Values in \textbf{bold} are the highest in the column. {*}Note that due to an imperfect balancing of data, chance level for humans (computed as majority class) is $0.124$.}
\label{tab:results}
\end{table}

\section{Results}
\label{sec:results}

\begin{figure*}[htb!]
\begin{center}
\includegraphics[width=0.74\textwidth]{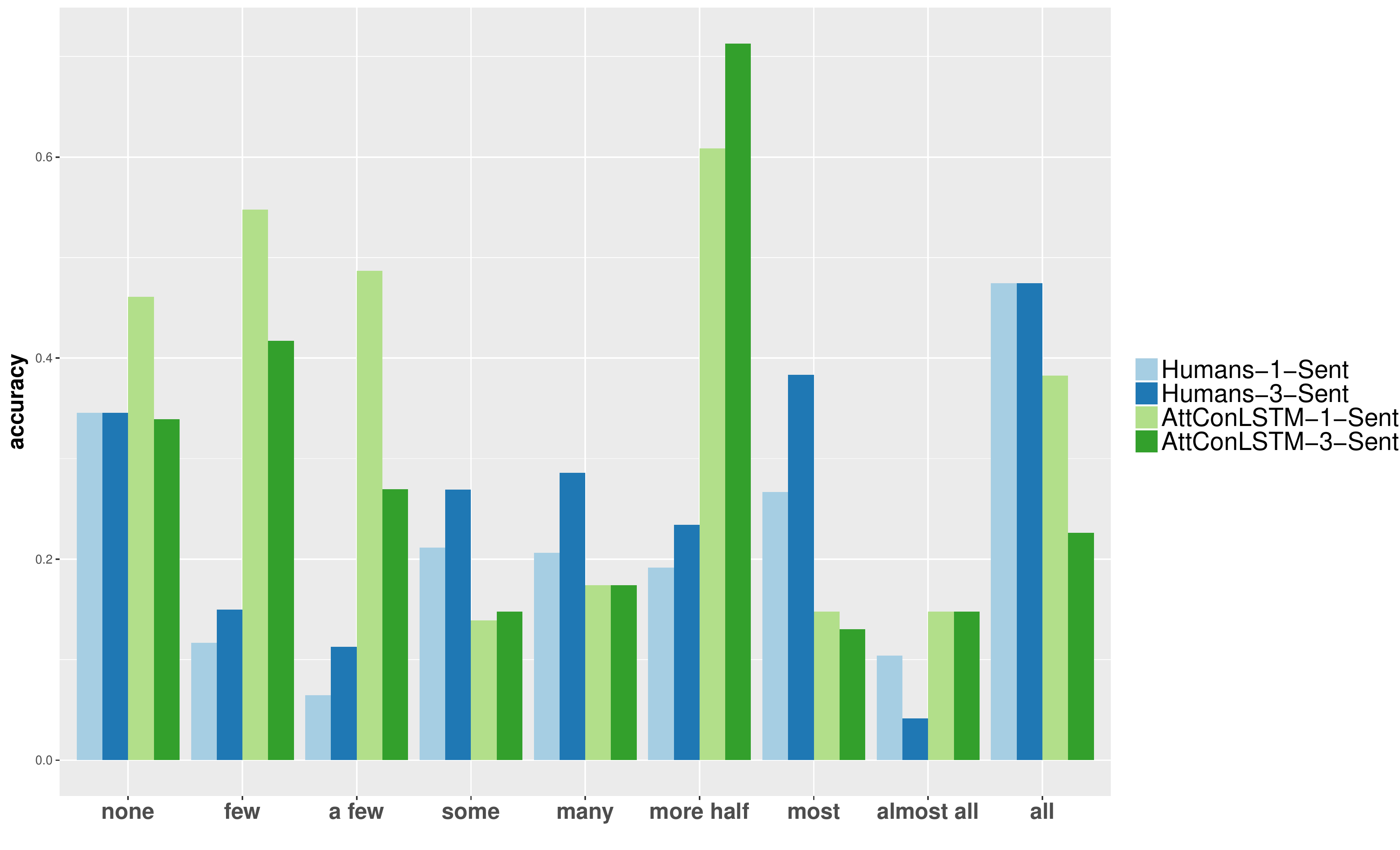}
\caption{Human vs \texttt{AttCon-LSTM} accuracy (\textit{val}) across quantifiers, loosely ordered by magnitude.}\label{fig:barplot}
\end{center}
\end{figure*}


Table~\ref{tab:results} reports the accuracy of all models and humans in both conditions. We have three main results. (1) Broader context \textit{helps} humans to perform the task, but \textit{hurts} model performance. This can be seen by comparing the 4-point increase of human accuracy from \texttt{1-Sent} (0.22) to \texttt{3-Sent} (0.26) with the generally worse performance of all models (e.g. \texttt{AttCon-LSTM}, from 0.34 to 0.27 in \textit{val}). (2) All models are significantly \textit{better} than humans in performing the task at the sentence level (\texttt{1-Sent}), whereas their performance is only slightly better than humans' in \texttt{3-Sent}. \texttt{AttCon-LSTM}, which is the best model in the former setting, achieves a significantly higher accuracy than humans' (0.34 vs 0.22). By contrast, in \texttt{3-Sent}, the performance of the best model is closer to that of humans (0.29 of \texttt{Att-LSTM} vs 0.26). It can be seen that LSTMs are overall the best-performing architectures, with \texttt{CNN} showing some potential in the handling of longer sequences (\texttt{3-Sent}). (3) As depicted in Figure~\ref{fig:barplot}, quantifiers that are easy/hard for humans are not necessarily easy/hard for the models. Compare `few', `a few', `more than half', `some', and `most': while the first three are generally hard for humans but predictable by the models, the last two show the opposite pattern. Moreover, quantifiers that are guessed by humans to a larger extent in \texttt{3-Sent} compared to \texttt{1-Sent}, thus profiting from the broader linguistic context, do not experience the same boost with models. Human accuracy improves notably for `few', `a few', `many', and `most', while model performance on the same quantifiers does not.

\begin{table}[b!]
\tiny
\centering
\begin{tabular}{|l|ccccccccc|}
\hline

\textit{none}           & \textbf{19} & 1          & 2 & 0           & 2           & 0          & 0           & 0          & 12          \\
\textit{few}            & 5           & \textbf{9} & 2 & 6           & 5           & 0          & 3           & 0          & 2           \\
\textit{a few}          & 0           & 0          & 7  & \textbf{17}    & 9           & 0          & 4           & 0          & 4           \\
\textit{some}           & 0           & 0          & 3 & \textbf{14} & 5           & 0    & 4           & 0          & 3           \\
\textit{many}           & 0           & 1          & 0 & 3           & \textbf{18} & 0          & 3           & 0          & 7           \\
\textit{more than half} & 0           & 0          & 0 & 2           & 2           & \textbf{11} & 10          & 4          & 2           \\
\textit{most}           & 0           & 0          & 0 & 1           & 7           & 0          & \textbf{23} & 4          & 8           \\
\textit{almost all}     & 0           & 1          & 0 & 3           & 2           & 1          & \textbf{7}           & 2 & 6           \\
\textit{all}            & 0           & 0          & 2 & 1           & 5           & 0          & 4           & 3          & \textbf{28} \\ \hline\hline

\textit{none}           & \textbf{39} & 15          & 13          & 10 & 0           & 20          & 5  & 3 & 10          \\
\textit{few}            & 3           & \textbf{48} & 18          & 7  & 9           & 20          & 5  & 1 & 4           \\
\textit{a few}          & 7           & 13          & \textbf{31} & 18 & 5           & 15          & 12 & 8 & 6           \\
\textit{some}           & 5           & 18          & 16          & 17 & 16          & \textbf{19}    & 9  & 5 & 10          \\
\textit{many}           & 2           & 18          & 18          & 15 & \textbf{20} & 17          & 10 & 6 & 9           \\
\textit{more than half} & 2           & 7           & 2           & 3  & 10          & \textbf{82} & 2  & 1 & 6           \\
\textit{most}           & 8           & 14          & 14          & 12 & 12          & \textbf{26}    & 15 & 5 & 9           \\
\textit{almost all}     & 5           & 9           & 15          & 10 & 8           & \textbf{37}    & 15 & 6 & 10          \\
\textit{all}            & 7           & 12          & 10          & 15 & 21          & 13          & 7  & 4 & \textbf{26} \\ \hline
\end{tabular}
\caption{Responses by humans (top) and \texttt{AttCon-LSTM} (bottom) in \texttt{3-Sent} (\textit{val}). Values in \textbf{bold} are the highest in the row.}\label{tab:errors}
\end{table}

To check whether humans and the models make similar errors, we look into the distribution of responses in \texttt{3-Sent} (\textit{val}), which is the most comparable setting with respect to accuracy. Table~\ref{tab:errors} reports responses by humans (top) and \texttt{AttCon-LSTM} (bottom). Human errors generally involve quantifiers that display a similar magnitude as the correct one. To illustrate, `some' is chosen in place of `a few', and `most' in place of either `almost all' or `more than half'. A similar pattern is observed in the model's predictions, though we note a bias toward `more than half'.




One last question concerns the types of linguistic cues exploited by the model (see section~\ref{sec:linganalysis}). We consider those cases which are correctly guessed by both humans and \texttt{AttCon-LSTM} in each setting and analyze the distribution of annotated cues. Non-semantic cues turn out to account for 41\% of cases in \texttt{3-Sent} and for 50\% cases in \texttt{1-Sent}. This analysis suggests that, compared to humans, the model capitalizes more on lexical, morpho-syntactic cues rather than exploiting the meaning of the context.

\section{Discussion}
\label{sec:discussion}

This study explored the role of linguistic context in predicting quantifiers. For humans, the task becomes easier when a broader context is given. For the best-performing LSTMs, broader context hurts performance. This pattern mirrors evidence that predictions by these models are mainly based on local contexts~\cite{hill:thego16}. Corroborating our hypotheses, \emph{proportional} quantifiers (`few', `many', `most') are predicted by humans with a higher accuracy with a broader context, whereas \textit{logical} quantifiers (`all', `none') do not experience a similar boost. Interestingly, humans are almost always able to grasp the magnitude of the missing quantifier, even when guessing the wrong one. This finding is in line with the overlapping meaning and use of these expressions~\cite{Moxey:1993zr}. It also provides indirect evidence for an ordered mental scale of quantifiers~\cite{holyoak1978,routh1994,moxey2000}. The reason why the models fail with certain quantifiers and not others is yet not clear. It may be that part of the disadvantage in the broader context condition is due to engineering issues, as suggested by an anonymous reviewer. We leave investigating these issues to future work. 







\section*{Acknowledgments}
We thank Marco Baroni, Raquel Fern{\'a}ndez, Germ{\'a}n Kruszewski, and Nghia The Pham for their valuable feedback. We thank the NVIDIA Corporation for the donation of GPUs used for this research, and the iV\&L Net (ICT COST Action IC1307) for funding the first author's research visit. This project has received funding from the European Research Council (ERC) under the European Union’s Horizon 2020 research and innovation programme (grant agreement No 716230).

\bibliography{paper}
\bibliographystyle{acl_natbib}




\end{document}